\documentclass{article} 
\usepackage{iclr2023/iclr2023_conference,times}

\usepackage{amsmath,amsfonts,bm}

\def\eqref#1{equation~\ref{#1}}

\def\1{\bm{1}}

\def\ra{{\textnormal{a}}}

\def\rva{{\mathbf{a}}}

\def\erva{{\textnormal{a}}}

\def\ervx{{\textnormal{x}}}

\def\rmA{{\mathbf{A}}}

\def\va{{\bm{a}}}

\def\ve{{\bm{e}}}

\def\eva{{a}}

\def\mA{{\bm{A}}}

\def\mI{{\bm{I}}}

\DeclareMathAlphabet{\mathsfit}{\encodingdefault}{\sfdefault}{m}{sl}
\SetMathAlphabet{\mathsfit}{bold}{\encodingdefault}{\sfdefault}{bx}{n}
\newcommand{\tens}[1]{\bm{\mathsfit{#1}}}
\def\tA{{\tens{A}}}

\def\gG{{\mathcal{G}}}

\def\sA{{\mathbb{A}}}
\def\sB{{\mathbb{B}}}

\def\emA{{A}}

\newcommand{\etens}[1]{\mathsfit{#1}}

\def\etA{{\etens{A}}}

\newcommand{\R}{\mathbb{R}}

\newcommand{\parents}{Pa} 

\DeclareMathOperator*{\argmax}{arg\,max}
\DeclareMathOperator*{\argmin}{arg\,min}

\usepackage[pagebackref=true,breaklinks=true,colorlinks,citecolor=brown]{hyperref}

\usepackage{url}
\usepackage{comment}
\usepackage{makecell, multirow}
\usepackage{multirow}
\usepackage{booktabs}
\usepackage{amsmath}
\usepackage{array}
\usepackage{mathrsfs}
\usepackage{amssymb}
\usepackage{amsfonts}
\usepackage{amsmath}
\usepackage[flushleft]{threeparttable}
\usepackage{tablefootnote}
\usepackage{multirow}
\usepackage{hhline}
\usepackage{graphicx}
\usepackage{graphicx, subcaption}
\usepackage{hanging}
\usepackage{color}
\usepackage{tikz}
\usepackage{float,wrapfig}
\usetikzlibrary{calc,arrows,decorations.markings}
\usepackage{balance}
\usepackage{hhline}
\usepackage{algorithm}
\usepackage{algpseudocode} 
\usepackage{ifthen}
\usepackage{xspace}

\usepackage{enumitem}

\makeatletter
\newcommand{\printfnsymbol}[1]{%
  \textsuperscript{\@fnsymbol{#1}}%
}

\newcommand{\ycReb}[1]{\textcolor{black}{{#1}}}

\newcommand{\hongyi}[1]{\textcolor{black}{#1}}

\newcommand{\netName}{LEAP\xspace}

\newcommand{\at}{\mathbf{a}_t}

\newcommand{\bs}{\mathbf{s}}
\newcommand{\ba}{\mathbf{a}}

\newcommand{\btau}[1]{\bm{\tau}^{#1}}

\newcommand\undermat[2]{%
  \makebox[0pt][l]{$\smash{\underbrace{\phantom{%
    \begin{matrix}#2\end{matrix}}}_{\text{$#1$}}}$}#2}

\newcommand{\expect}[2]{\mathbb{E}_{#1} \left[ #2 \right] }
\newcommand{\te}[1]{\texttt{#1}}

\newcommand{\Hrz}{T}
\newcommand{\DataNum}{M}
\newcommand{\IterNum}{N}

\newcommand{\enModel}{E}
\newcommand{\mParam}{\theta}

\newcommand{\sect}[1]{Section~\ref{#1}}

\newcommand{\myparagraph}[1]{\vspace{-8pt}\paragraph{#1}}

\definecolor{MyDarkBlue}{rgb}{0,0.08,1}
\definecolor{MyDarkGreen}{rgb}{0.02,0.6,0.02}
\definecolor{MyDarkRed}{rgb}{0.8,0.02,0.02}
\definecolor{MyDarkOrange}{rgb}{0.40,0.2,0.02}
\definecolor{MyPurple}{RGB}{111,0,255}
\definecolor{MyRed}{rgb}{1.0,0.0,0.0}
\definecolor{MyGold}{rgb}{0.75,0.6,0.12}
\definecolor{MyDarkgray}{rgb}{0.66, 0.66, 0.66}

\newcommand{\approptoinn}[2]{\mathrel{\vcenter{
  \offinterlineskip\halign{\hfil$##$\cr
    #1\propto\cr\noalign{\kern2pt}#1\sim\cr\noalign{\kern-2pt}}}}}

\title{Planning with \hongyi{Sequence} Models through Iterative Energy Minimization}

\author{
\qquad Hongyi Chen\thanks{\fontsize{6}{7} denotes equal contribution. Correspondence to \texttt{hchen657@gatech.edu},\texttt{ yilundu@mit.edu}, \texttt{yychen2019@gatech.edu}}\;\;\printfnsymbol{2}, ~Yilun Du \printfnsymbol{1}\printfnsymbol{3}, ~Yiye Chen\printfnsymbol{1}\printfnsymbol{2}, ~Joshua Tenenbaum\printfnsymbol{3}, ~Patricio Vela\printfnsymbol{2}\\
\qquad\qquad~Georgia Institute of Technology\printfnsymbol{2}\qquad ~Massachusetts Institute of Technology\printfnsymbol{3}\\
}

\begin{document}

\maketitle

\begin{abstract}
Recent works have shown that \hongyi{sequence} modeling can be effectively used to train reinforcement learning (RL) policies.  
However, the success of applying existing sequence models to planning, 
in which we wish to obtain a trajectory of actions to reach some goal,
is less straightforward. 
The typical autoregressive generation procedures of \hongyi{sequence} models preclude sequential refinement of earlier steps, which limits the effectiveness of a predicted plan. 
In this paper, we suggest an approach towards integrating planning with \hongyi{sequence} models based on the idea of iterative energy minimization, and illustrate how such a procedure leads to improved RL performance across different tasks.
We train a masked language model to capture an implicit energy function over trajectories of actions, 
and formulate planning as finding a trajectory of actions with minimum energy. 
We illustrate how this procedure enables improved performance over recent approaches across BabyAI and Atari environments. 
We further demonstrate unique benefits of our iterative optimization procedure,
involving new task generalization, test-time constraints adaptation, and the ability to compose plans together. Project website: \href{https://hychen-naza.github.io/projects/LEAP/index.html}{https://hychen-naza.github.io/projects/LEAP}.






\end{abstract}

\section{Introduction}
\begin{wrapfigure}{r}{0.5\textwidth}
   \vspace{-10pt}
  \centering
  \scalebox{0.97}{
    \begin{tikzpicture}
     \node[anchor=north west](graph) at (0in,0in)
      {{\includegraphics[width=0.5\textwidth,trim={0.1in
      8.2in 2.2in 0.0in},clip=true]{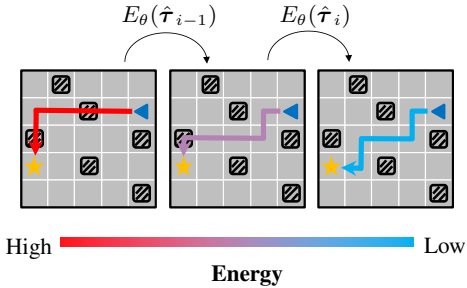}}};
    \node[anchor=south] at ($ (graph.west) + (2.2, 1.3)$)  {\small $\enModel_\mParam(\hat{\btau{}}_{i-1})$};
    \node[anchor=south] at ($ (graph.west) + (4.2, 1.3)$)  {\small $\enModel_\mParam(\hat{\btau{}}_{i})$};
     \node[anchor=north] at ($ (graph.south) + (-0.35, 0.0) $) {\bf \small Energy};
    \node[anchor=east] at ($ (graph.south) + (-2.9, 0.1) $) { \footnotesize High};
     \node[anchor=west] at ($ (graph.south) + (1.95, 0.15) $) { \footnotesize Low};
    \end{tikzpicture}
  }
  \vspace{-20pt}
  \caption{\small
  \textbf{Plan Generation through Iteratively Energy Minimization.}
  \netName plans a trajectory to a goal (specified by the yellow star) by iteratively sampling and minimizing a trajectory energy function estimated using language model $E_\theta$.
  }
 \label{fig:teaser}
 \vspace{-15pt}
\end{wrapfigure}

Sequence modeling has emerged as unified paradigm to study numerous domains such as language \citep{brown2020gpt3,radford2018improving} and vision \citep{yu2022scaling,dosovitskiy2020image}. Recently,  \citep{chen2021dt,janner2021tt} have shown how a similar approach can be effectively applied to decision making, by predicting the next action to take. However, in many decision making domains, it is sub-optimal to simply predict the next action to execute -- as such an action may be only locally optimal and lead to global dead-end. Instead, it is more desirable to plan a sequence of actions towards a final goal, and choose the action most optimal for the final overall goal. 

Unlike greedily picking  the next action to execute, effectively constructing an action sequence towards a given goal requires a careful, iterative procedure, where we need to assess and refine intermediate actions in a plan to ensure we reach the final goal. To refine an action at a particular timestep in a plan, we must reconsider both actions both before and after the chosen action. Directly applying this procedure to standard language generation is difficult, as the standard autoregressive decoding procedure prevents regeneration of previous actions based of future ones. For example, if the first five predicted actions places an agent at a location too far to reach a given goal, there is no manner we may change the early portions of plan.

In this paper, we propose an approach to iteratively generate plans using \hongyi{sequence} models. Our approach, Mu\textbf{l}tistep \textbf{E}nergy-Minimiz\textbf{a}tion \textbf{P}lanner (\netName), formulates planning as an iterative optimization procedure on an energy function over trajectories defined implicitly by a \hongyi{sequence} model (illustrated in Figure~\ref{fig:teaser}). To define an energy function across trajectories, we train a bidirectional \hongyi{sequence} model using a masked-language modeling (MLM) objective \citep{devlin-etal-2019-bert}. We define the energy of a trajectory as the negative pseudo-likelihood (PLL) of this MLM \citep{salazar2019masked} and sequentially minimize this energy value by replacing actions at different timepoints in the trajectory with the marginal estimates given by the MLM. Since our MLM is bi-directional in nature, the choice of new action at a given time-step is generated based on both future and past actions.

By iteratively generating actions through planning, we illustrate how our proposed framework outperforms prior methods in both BabyAI \citep{babyai_iclr19} and  Atari \citep{bellemare2013arcade} tasks. Furthermore, by formulating the action generation process as an iterative energy minimization procedure, we illustrate how this enables us to generalize to environments with new sets of test-time constraints as well as more complex planning problems. Finally, we demonstrate how such an energy minimization procedure enables us to compose planning procedures in different models together, enabling the construction of plan which achieves multiple objectives.

Concretely, in this paper, we contribute the following: First, we introduce \netName, a framework through which we may iteratively plan with \hongyi{sequence} models. Second, we illustrate how such a planning framework can be beneficial on both BabyAI and Atari domains. Finally, we illustrate how iteratively planning through energy minimization gives a set of unique properties, enabling better test time performance on more complex environments and environments with new test-time obstacles, and the ability to compose multiple learned models together, to jointly generate plans that satisfy multiple sets of goals.

\vspace{-5pt}
\section{Related Work}
\vspace{3pt}
\myparagraph{\hongyi{Sequence} Models and Reinforcement Learning.} Sequence modeling with deep networks, from sequence-to-sequence models \citep{LSTM1997,sutskever2014sequence} to BERT \citep{devlin-etal-2019-bert} and XLnet \citep{NEURIPS2019_dc6a7e65}, have shown promising results in a series of language modeling problems \citep{dai2019transformer,sutskever2014sequence,liu2019text, dehghani2018universal}. With these advances, people start applying sequence models to represent components in standard RL such as policies, value functions, and models to improved performance \citep{rl2018lstm,RL2020Transformer,kapturowski2018RLrecurrent}. While the sequence models provide memory information to make the agent predictions temporally and spatially coherent, they still rely on standard RL algorithm to fit value functions or compute policy gradients. Furthermore, recent works replace as much of the RL pipeline as possible with sequence modeling to leverage its scalability, flexible representations and causally reasoning \citep{janner2021tt, chen2021dt,furuta2021generalized, zheng2022online,emmons2021rvs,li2022pre}. However, those methods adopt autoregressive modeling objectives and the predicted trajectory sequences have no easy way to be optimized, which will inevitably lower the long-horizon accuracy. \hongyi{Recent studies point out that using sequence models \citep{chen2021dt, emmons2021rvs} rather than typical value-based approaches have difficulty converging in stochastic environments \citep{paster2020planning, dtStoch}.}


\myparagraph{Planning in Reinforcement Learning.} Planning has been explored extensively in model-based RL, which learns how the environment respond to actions \citep{sutton1991dyna}. The learned world dynamic model is exploited to predict the conditional distribution over the immediate next state or autoregressively reason the long-term future \citep{chiappa2017recurrent_env,ke2018modeling}. However, due to error compounding, plans generated by this procedure often look more like adversarial examples than optimal trajectories when the planning horizon is extended \citep{bengio2015scheduled, talvitie2014model, asadi2018lipschitz}. To avoid the aforementioned issues, simple gradient-free method like Monte Carlo tree search \citep{10.1007/978-3-540-75538-8_7}, random shooting \citep{nagabandi2018neural} and \hongyi{beam search \citep{plate}} are explored. Another line of works studied how to break the barrier between model learning and planning, and plan with an imperfect model, include training an autoregressive latent-space model to predict values for abstract future states \citep{tamar2016value, oh2017value, schrittwieser2020mastering, plate}; energy-based models of policies for model-free reinforcement learning \citep{haarnoja2017reinforcement}; \hongyi{improve the offline policies by planning with learned models for model-based reinforcement learning \citep{mopo, muzeroUnp}}; directly applying collocation techniques for direct trajectory optimization \citep{erez2012trajectory, du2019model}; and folding planning into the generative modeling process \citep{janner2021tt}. In contrast to these works, we explore having planning directly integrated in a language modeling framework.

\myparagraph{Energy Based Learning.} Energy-Based Models (EBMs) capture dependencies between variables by associating a scalar energy to each configuration of the variables, and provide a unified theoretical framework for many probabilistic and
non-probabilistic approaches to learning \citep{lecun2006tutorial}. Prior works have explored EBMs for policy training in model-free RL \citep{haarnoja2017reinforcement}, modeling the environment dynamics in model-based RL \citep{du2019model, janner2022planning} and natural images \citep{du2019implicit, dai2019exponential}, as well as energy values over text \citep{goyal2021exposing}. Most similar to our work, \citet{du2019model} illustrates how energy optimization in EBMs naturally support planning given start and goal state distributions. However, the 
 underlying training relies on a constrastive divergence, which is difficult to train \citep{du2020improved}. 
 On the other hand, our training approach relies on a more stable masked language modeling objective.

\vspace{-8pt}
\section{Method}
\vspace{-5pt}
\label{method}
In this section, we describe our framework, Mu\textbf{l}tistep \textbf{E}nergy-Minimiz\textbf{a}tion \textbf{P}lanner (\netName), which formulates planning as a energy minimization procedure. Given a set of trajectories in a discrete action space, with each trajectory containing state and action sequences $(\bs_1, \ba_1, \bs_2, \ba_2, \ldots, \bs_N, \ba_N)$, our goal is to learn a planning model which can predict a sequence of actions $\ba_{1:\Hrz}$, given the \hongyi{trajectory context $\btau{}_{ctx}$ containing the past $K$ steps states and actions}, that maximizes the long-term task-dependent objective $\mathcal{J}$:
\vspace{-3pt}
\[
\ba_{1:\Hrz}^* = \argmax_{\ba_{1:\Hrz}} \mathcal{J}(\hongyi{\btau{}_{ctx}}, \ba_{1:\Hrz})
\]
where \hongyi{$N$ denotes the length of the entire trajectory and} $\Hrz$ is the planning horizon. We use the abbreviation $\mathcal{J}(\btau{})$, \hongyi{where $\btau{}=:(\btau{}_{ctx}, \ba_{1:\Hrz})$}, to denote the objective value of that trajectory. To formulate this planning procedure, we learn an energy function $E_\theta(\btau{})$, which maps each trajectory $\btau{}$ to a scalar valued energy so that
\vspace{-3pt}
\[
\ba_{1:\Hrz}^* = \argmin_{\ba_{1:\Hrz}} E_\theta(\hongyi{\btau{}_{ctx}}, \ba_{1:\Hrz}).
\]

\vspace{-10pt}
\subsection{Learning Trajectory Level Energy Functions}

We wish to construct an energy function $E_\theta(\hongyi{\btau{}_{ctx}}, \ba_{1:\Hrz})$ such that minimal energy is assigned to optimal set actions $\ba_{1:\Hrz}^*$. To train our energy function, we assume access to dataset of $\DataNum$ near optimal set of demonstrations in the environment, and train our energy function to have low energy across demonstrations. Below, we introduce masked language models, and then discuss how we may get such a desired energy function from masked language modeling.

\myparagraph{Masked Language Models.} Given a trajectory of the form $(\bs_1, \ba_1, \bs_2, \ba_2, \ldots, \bs_n, \ba_n)$, we train a transformer language model to model the marginal likelihood $p_\theta(\ba_t | \hongyi{\btau{}_{ctx}}, \ba_{-t})$, where we utilize $\ba_{-t}$ as shorthand for actions \hongyi{$\ba_{1:\Hrz}$} except the action at timestep $t$. To train this masked language model (MLM), we utilize the standard BERT training objective \citep{devlin-etal-2019-bert}, where we minimize the loss function
\begin{equation}
    \mathcal{L}_{\text{MLM}} = \expect{\btau{}, t}{-\log p_\theta(\ba_t;\hongyi{\btau{}_{ctx}}, \ba_{-t})},
\end{equation}
where we mask out and predict the marginal likelihood of percentage of the actions in a trajectory (details on masking in the \sect{appendix:experiment}). 
.

\begin{figure*}[t!]
 
  \vspace*{-0.0in}
  \centering
  \includegraphics[width=0.9\textwidth]{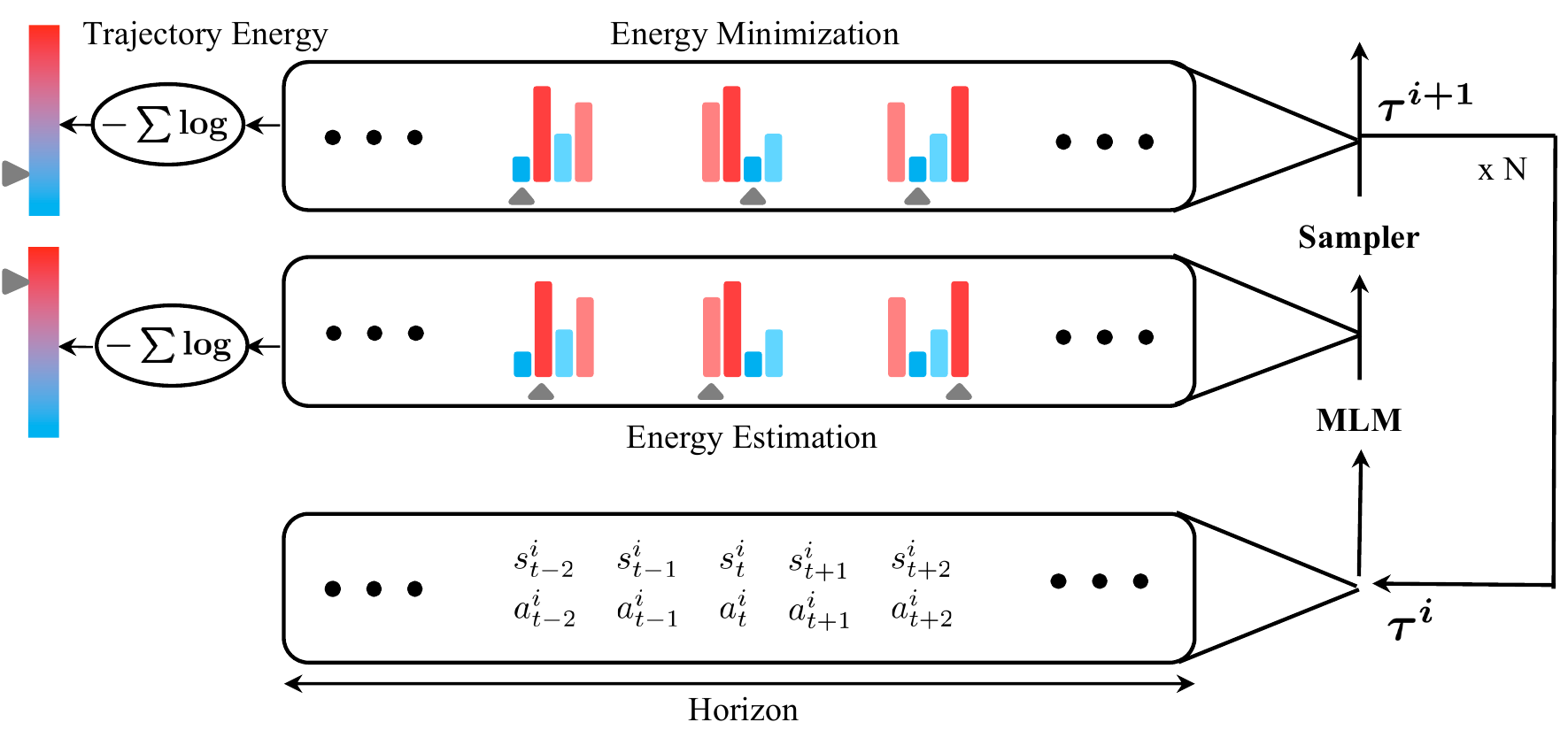}
  \vspace{-10pt}
  \caption{\small \textbf{Energy Minimization in \netName.}
    \netName generates plans via Gibbs sampling different actions based on a learned trajectory energy model $E_\theta(\btau{})$.
    In each iteration,  Masked Language Model (MLM) predicts the energy of alternative actions at selected timesteps in a trajectory.
    A new trajectory is generated using a Gibbs sampler, with individual actions sampled based on the energy distribution.
    By repeating the above steps iteratively, \netName generates the trajectory with low energy value.
  }
  \vspace{-10pt}
 \label{fig:method}
\end{figure*}

\myparagraph{Constructing Trajectory Level Energy Functions.} Given a trained MLM, we define an energy function for a trajectory as the sum of negative  marginal likelihood of each action in a sequence
\begin{equation}
    E_\theta(\btau{}) = -\sum_t \log p_\theta(\ba_t;\hongyi{\btau{}_{ctx}}, \ba_{-t}).
    \label{eqn:energy}
\end{equation}
Such an energy function, also known as the pseudolikelihood of the MLM, has been used extensively in prior work in NLP \citep{goyal2021exposing, salazar-etal-2020-masked}, and has been demonstrated to effectively score the quality natural language text (outperforming direct autoregressive scoring) \citep{salazar-etal-2020-masked}. In our planning context, this translates to effectively assigning low energy to optimal planned actions, which is our desired goal for $E_\theta(\btau{})$. We illustrate the energy computation process in Figure~\ref{fig:method}.

\vspace{-3pt}
\subsection{Planning as Energy Minimization}
\vspace{-3pt}

Given a learned energy function $E_\theta(\btau{})$ defined through Equation~\ref{eqn:energy}, which assigns low energy to optimal trajectories \hongyi{$\ba_{1:\Hrz}^*$}, we wish to plan a sequence of actions at test-time that minimizes our energy function. To implement this planning procedure, we utilize Gibbs sampling across actions in the trajectory to iteratively refine and construct a trajectory $\btau{}$ that minimizes $E_\theta(\btau{})$.

In particular, for test-time plan generation, we initialize a masked trajectory of length $\Hrz$ with a small context length of past states and actions, which we illustrate in Equation~\ref{eq:traj_with_pad_actions}. At each step of Gibb's sampling, we randomly mask out one or multiple action tokens and use the MLM to estimate energy distribution across actions at each masked token. 
Then, action $\at$ and each token position is sampled using Gibb's sampling based on the locally normalized energy score of each token $\at \sim p_\theta(\ba_t;\hongyi{\btau{}_{ctx}}, \ba_{-t})$. 
The process is illustrated in Figure~\ref{fig:method}, where the actions with low energy values (in blue) are sampled to minimize $E_\theta(\btau{i};\theta)$ in each iteration. 
To sample effectively from the underlying energy distribution, we repeat this procedure for multiple timesteps, which we illustrate in  Algorithm~\ref{algo:iterative} which aims to minimize the overall energy expression in Equation~\ref{eqn:energy}.
\begin{equation}
\btau{} = 
\begin{bmatrix}
\bs_1 & \bs_2 & \dots & \bs_{n-1} & \bs_n & \bs_n & \dots & \bs_n\\
\undermat{\text{context}}{\ba_1 & \ba_2 & \dots & \ba_{n-1}} & \undermat{\text{plan}}{\left[ \texttt{PAD} \right] & \left[ \texttt{PAD} \right] & \dots & \left[ \texttt{PAD} \right]}
\end{bmatrix}
\label{eq:traj_with_pad_actions}
\end{equation}
\vspace{-10pt}
%
\begin{figure}[h]
\centering
\scalebox{0.75}{
\begin{minipage}{\linewidth}
\begin{algorithm}[H]
	\caption{Iterative Planning through Energy Minimization (for discrete actions)} 
	\label{algo:iterative}
	\begin{algorithmic}[1]
	\State \textbf{Require} trained energy model $h_\theta$, context trajectory $\bm{\tau}$
	\State Pad the states and actions with length $T$ into context trajectory 
    \For {$i=1,\ldots, \IterNum$}
        \State
        \small{\color{gray} 
        \te{// Sample index set}}.
        \State 
        $I \sim [1, 2, \cdots, \Hrz]$
        \State \small{\color{gray}\te{// Estimate the energy distributions on masked tokens}}
        \State $\mathcal{E}$ $\leftarrow$ $f(h(\btau{i}_{\backslash I}; \theta))$
        \State \small{\color{gray}\te{// Sample the action tokens based on energy value}} 
        \State $\ba \sim \mathcal{E}$ 
        \State 
        \small{\color{gray} \te{// Update actions $\ba$ in $\bm{\tau}$ at masked tokens}}
        \State
        $\btau{i+1} \leftarrow \btau{i}_{\backslash \Hrz} + \ba$
    \EndFor
    \State Execute all planned actions $\ba_{1:T}$ or the first planned action $\ba_1$ in padded trajectory $\bm{\tau}$
	\end{algorithmic} 
\end{algorithm}
\end{minipage}
}
\end{figure}
\vspace{-5pt}

Note that our resultant algorithm has several important differences compared to sequential model based approaches such as Decision Transformer (DT) \citep{chen2021dt}. First, actions are generated using an energy function defined globally across actions, enabling us to choose each action factoring in the entire trajectory context. Second, our action generation procedure is iterative in nature, allowing us to leverage computational time to find better solutions towards final goals.

\vspace{-5pt}
\section{Properties of Multistep Energy-Minimization Planner} 
\vspace{-3pt}
\label{sec:prop}

\begin{figure*}[t!]
 
  \vspace*{-0.0in}
  \centering
  \scalebox{0.90}{
    \begin{tikzpicture}
     \node[anchor=north west] at (0in,0in)
      {{\includegraphics[width=1.0\textwidth,clip=true,trim=0in
      6.8in 0.3in 0.0in]{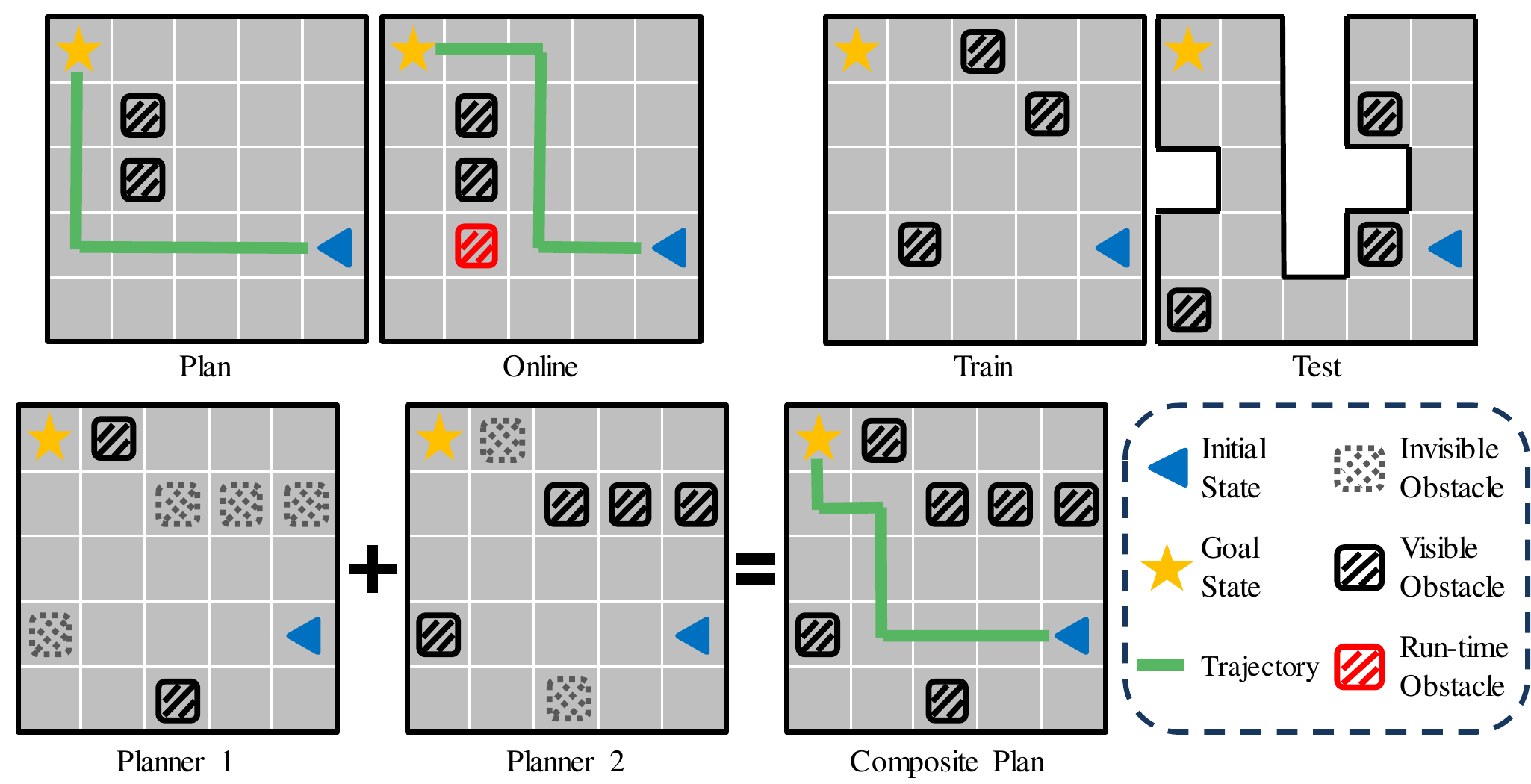}}};
    \node[yshift=-0pt,anchor=north west] at (-0.05in,-0.08in) {\bf (a)};
    \node[yshift=-0pt,anchor=north west] at (2.72in,-0.08in) {\bf (b)};
    \node[yshift=-0pt,anchor=north west] at (-0.15in,-1.47in) {\bf (c)};
    \end{tikzpicture}
  }
  \vspace{-10pt}
  \caption{\small
    \textbf{Properties of Planning as Energy Minimization.} By formulating planning as energy minimization, \netName enables the following properties:
    (a): Online adaptation;
    (b): Generalization;
    (c): Task composition.
  }
  \vspace{-15pt}
 \label{fig:property_method}
\end{figure*}

In \netName, we formulate planning as an optimization procedure $\argmin_{\btau{}} E_\theta(\btau{})$. 
By formulating planning in such a manner, we illustrate how
\hongyi{our approach enables} flexible online adaptation to new test-time constraints, generalization to harder planning problems, and plan composition to achieve multiple set of goals (illustrated in Figure~\ref{fig:property_method}).

\myparagraph{Online adaptation.} In \netName, recall that plans are generated by minimizing an energy function $E_{\theta}(\btau{})$ across trajectories. At test time, if we have new external constraints, we maybe correspondingly define a new energy function $E_{\text{constraint}}(\btau{})$ to encode these constraints. For instance, if a state becomes dangerous at test time (illustrated as a red grid in Figure~\ref{fig:property_method} (a)) -- we may directly define an energy function which assigns 0 energy to plans which do not utilize this state and high energy to plans which utilize such a state. We may then generate plans which satisfies this constraint by simply minimizing the summed energy function 
\begin{equation}
   \bm{\tau}^{*} = \argmin_{\btau{}} (E_\theta(\btau{}) + E_{\text{constraint}}(\btau{})).
\end{equation}
and then utilize Gibb's sampling to obtain a plan $\bm{\tau}^{*}$ from $E_\theta(\btau{}) + E_{\text{constraint}}(\btau{})$. \hongyi{While online adaptation may also be integrated with other sampling based planners using rejection sampling,   our approach directly integrates trajectory generation with online constraints.}




\myparagraph{Novel Environment Generalization.} In \netName, we leverage many steps of sampling to recover an optimal trajectory $\bm{\tau}^{*}$ which minimizes our learned trajectory energy function $E_{\theta}(\btau{})$. In settings at test time when the environment is  more complex than those seen at training time (illustrated in Figure~\ref{fig:property_method} (b)) -- the underlying energy function necessary to compute plan feasibility may remain simple (i.e. measure if actions enter obstacles), but the underlying planning problem becomes much more difficult. In these settings, as long as the learned function $E_{\theta}(\btau{})$ generalizes, and we may simply leverage more steps of sampling to recover a successful plan in this more complex environment.



\myparagraph{Task compositionality.} Given two different instances of \netName, $E_\theta^1(\btau{})$ and $E_\theta^2(\btau{})$, encoding separate tasks for planning, we may generate a trajectory which accomplishes the tasks encoded by both models by simply minimizing a composed energy function \hongyi{(assuming task independence)} 
\begin{equation}
   \bm{\tau}^{*} = \argmin_{\btau{}} (E_\theta^1(\btau{}) + E_\theta^2(\btau{})).
\end{equation}
An simple instance of such a setting is illustrated in Figure~\ref{fig:property_method} (c), where the first \netName model $E_\theta^1(\btau{})$ encodes two obstacles in an environment, and a second \netName model  $E_\theta^2(\btau{})$ encodes four other obstacles. By jointly optimizing both energy functions (through Gibbs sampling), we may successfully construct a plan which avoids all obstacles in both models.

\section{Experiments}
\vspace{-3pt}

\begin{figure*}[t]
\centering
\includegraphics[width=0.6\textwidth]{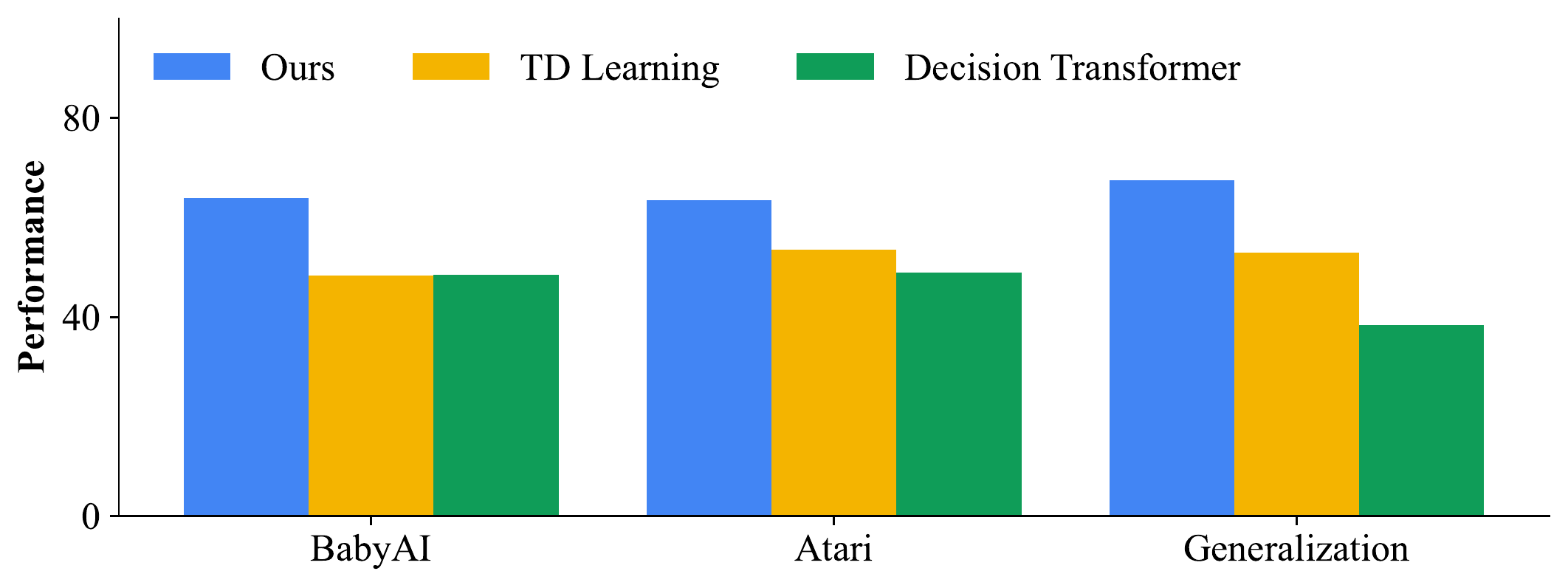}
\vspace{-10pt}
\caption{\small
\textbf{Quantitative Results of \netName of Different Domains.} Results comparing \netName to Decision Transformer and TD learning (IQL in BabyAI and Generalization tests and CQL in Atari) across BabyAI, Atari, and Generalization Tests. On a diverse set of tasks, \netName performs better than prior approaches.} 
\label{fig:main_results}
\vspace{-10pt}
\end{figure*}

In this section, we evaluate the planning performance of \netName in BabyAI and Atari environments. We compare with a variety of different offline reinforcement learning approaches, and summarize the main results in Figure~\ref{fig:main_results}.

\vspace{-3pt}
\subsection{BabyAI}
\vspace{3pt}
\myparagraph{Setup} The BabyAI comprises an extensible suite of tasks in environments with different sizes and shapes, where the reward is given only when the task is successfully finished. We evaluate models in trajectory planning that requires the agent move to the goal object through turn left, right and move forward actions, and general instruction completion tasks in which the agent needs to take extra pickup, drop and open actions to complete the instruction. For more detailed experimental settings and parameters, please refer to Table \ref{tbl:babyai_env_setting} in Appendix. 


\myparagraph{Baselines} 
We compare our approach with a set of different baselines: Behavior Cloning algorithm (BC); model free reinforcement learning (RL) algorithms Batch-Constrained deep Q-Learning (BCQ) \citep{fujimoto2019off}, and Implicit Q-Learning (IQL) \citep{kostrikov2021offline}; \hongyi{model based RL algorithm Model-based Offline Policy (MOPO) \citep{mopo}}; return-conditioning approach Decision Transformer (DT) \citep{chen2021dt}; \hongyi{model based Planning Transformer (PlaTe) \citep{plate}}. In our experiments, we use, as model inputs, the full observation of the environment, the instruction, the agent's current location and the goal object location (if available).

\myparagraph{Results}
Across all environments, \netName achieves highest success rate, with the margin magnified on larger, harder tasks, see Table~\ref{tbl:babyai_main}. In particular, \netName could solve the easy tasks like GoToLocalS8N7 with nearly 90\% success rate, and has huge advantages over baselines in the large maze worlds (GoToObjMazeS7) and complex tasks (GoToSeqS5R2) which require going to a sequence of objects in correct order. In contrast, baselines perform poorly solving these difficult tasks. 

\begin{figure*}[t!]
 
  \vspace*{-0.0in}
  \centering
  \scalebox{0.85}{
    \begin{tikzpicture}
     \node[anchor=north west](main) at (0in,0in)
      {{\includegraphics[width=0.9\textwidth,clip=true,trim=0in
      7.8in 0.0in 0.5in]{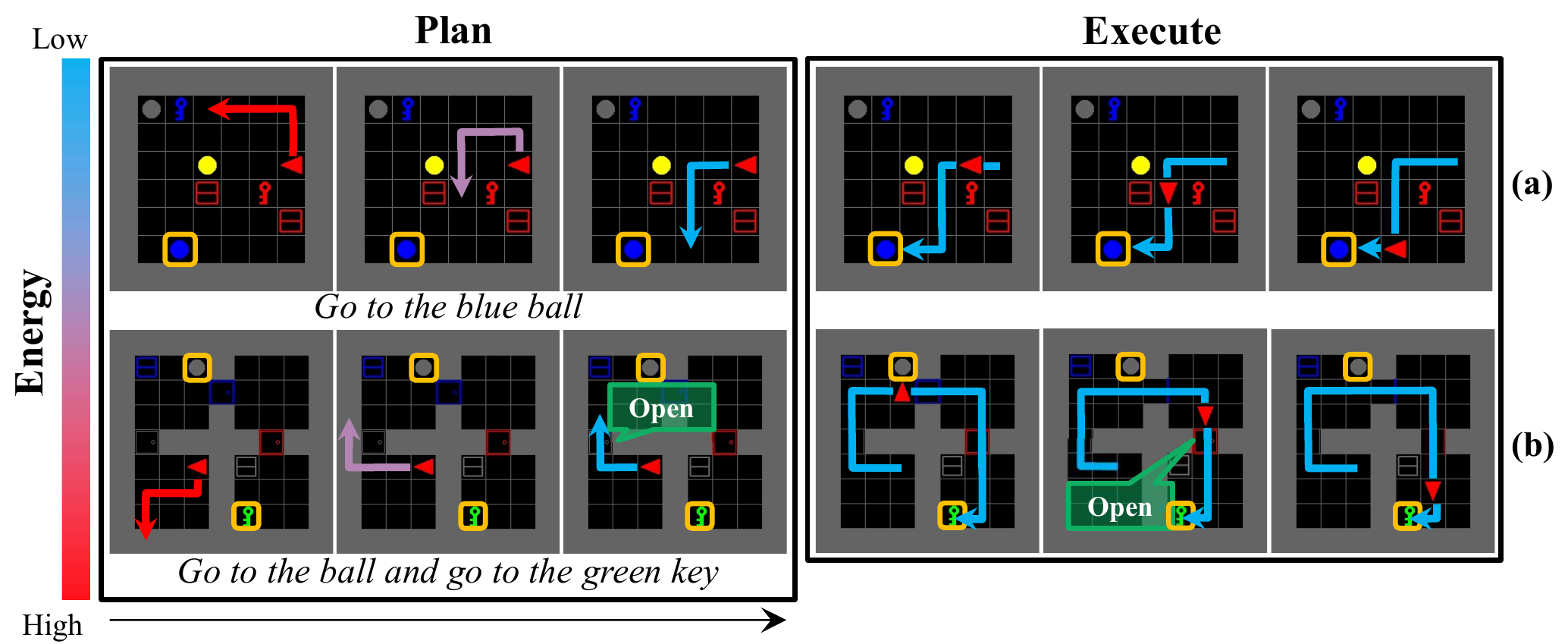}}};
     \node[anchor=north] at ($ (main.south) + (-4.45, 0.3) $){\footnotesize \IterNum=1};
     \node[anchor=north] at ($ (main.south) + (-2.65, 0.3) $){\footnotesize \IterNum=5};
     \node[anchor=north] at ($ (main.south) + (-0.85, 0.3) $){\footnotesize \IterNum=10};
    \end{tikzpicture}
  }
  \vspace{-8pt}
  \caption{
    \textbf{\small Qualitative Visualization of Planning and Execution Procedure in BabyAI}.
    \textit{Left} depicts the planning through iterative energy minimization where $N$ is the sample iteration number.
    \textit{Right} shows the execution of the concatenate action sequences.
    Two task settings are illustrated:
    \textbf{(a)}: \textbf{Trajectory planning}, where the task is to solely plan a trajectory leading to the goal location.
    \textbf{(b)}: \textbf{Instruction completion}, where a sequence of tasks are commanded, and an additional "Open" is involved to get through the doors.
    Target locations are marked with \protect{\raisebox{-.05cm}{\includegraphics[height=.30cm]{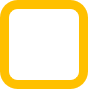}}}.
  }
  \vspace*{-10pt}
 \label{fig:babyai_plan_execute}
\end{figure*}

Next, we visualize the underlying iterative planing and execution procedure in GoToLocalS8N7 and GoToSeqS5R2 tasks. On the left side of Figure~\ref{fig:babyai_plan_execute}, we present how the trajectory is optimized and its energy is minimized at single time step. Through iterative refinement, the final blue trajectory is closer to the optimal solution than the original red one, which follows the correct direction with higher efficiency and perform the actions like open the door in correct situation. On the right side, we present the entire task completion process through many time steps. \netName successfully plans a efficient trajectory to visit the two objects and opens the door when blocked. \hongyi{We also explore the model performance in the stochastic settings, please refer to Appendix \ref{appendix:stochastic}.}

\begin{table*}[h]
\centering
\small
\scalebox{0.8}{
\begin{tabular}{@{}ccccccccc@{}}
\toprule
\multicolumn{1}{c}{\bf Task} & \multicolumn{1}{c}{\bf Env} & \multicolumn{1}{c}{\bf BC} & \multicolumn{1}{c}{\bf BCQ} & \multicolumn{1}{c}{\bf IQL} & \multicolumn{1}{c}{\bf DT} & \multicolumn{1}{c}{\bf \hongyi{PlaTe}} & \multicolumn{1}{c}{\bf \hongyi{MOPO}}& \multicolumn{1}{c}{\bf \netName} \\
 \midrule
\multirow{4}*{\makecell[c]{Trajectory\\Planning}} 
&{GoToLocalS7N5} & 71.0\% & 71.5\% & 84.5\%&73.0\% & \hongyi{42.5\%} & \hongyi{87.0\%}& \bf{91.0\%} \\
&{GoToLocalS8N7} & 61.5\%&63.0\% & 71.5\%&63.5\%&\hongyi{45.0\%} & \hongyi{81.5\%}&\bf{95.0\%}\\
&{GoToObjMazeS4} & 24.0\%&23.0\%&52.5\%&46.5\% & \hongyi{35.5\%} & \hongyi{60.0\%}& \bf{62.5\%}\\
&{GoToObjMazeS7} & 18.0\%&10.5\% & 29.0\%&22.0\%&\hongyi{27.5\%} & \hongyi{30.5\%}&\bf{42.5\%}\\\midrule
\multirow{3}*{\makecell[c]{Instruction\\ Completion}} 
&{PickUpLoc} & 57.5\% & 58.5\%& 41.0\%&59.5\% & \hongyi{7.5\%} &\hongyi{43.5\%}& \bf{67.0\%}\\
&{GoToSeqS5R2} & 13.5\% & 10.0\%& 28.5\%&26.5\% & \hongyi{25.0\%}&\hongyi{30.0\%}& \bf{33.0\%} \\
&{GoToObjMazeS4R2Close} & 24.0\%&22.5\%&31.5\% &48.5\% & \hongyi{32.5\%}& \hongyi{36.0\%}& \bf{55.0\%}\\
\bottomrule
\end{tabular}
}
\caption{\small \textbf{BabyAI Quantitative Performance.} The task success rate of \netName and a variety of prior algorithms on BabyAI env. Models are trained with 500 optimal trajectory demos in each environment, and results are averaged over 5 random seeds. The SX and NY in environment name means its size and the number of obstacles.}
\label{tbl:babyai_main}
\vspace{-10pt}
\end{table*}

\begin{figure*}[t!]
    \centering
    \begin{minipage}{0.43\textwidth}
        \centering
        \includegraphics[width=\textwidth]{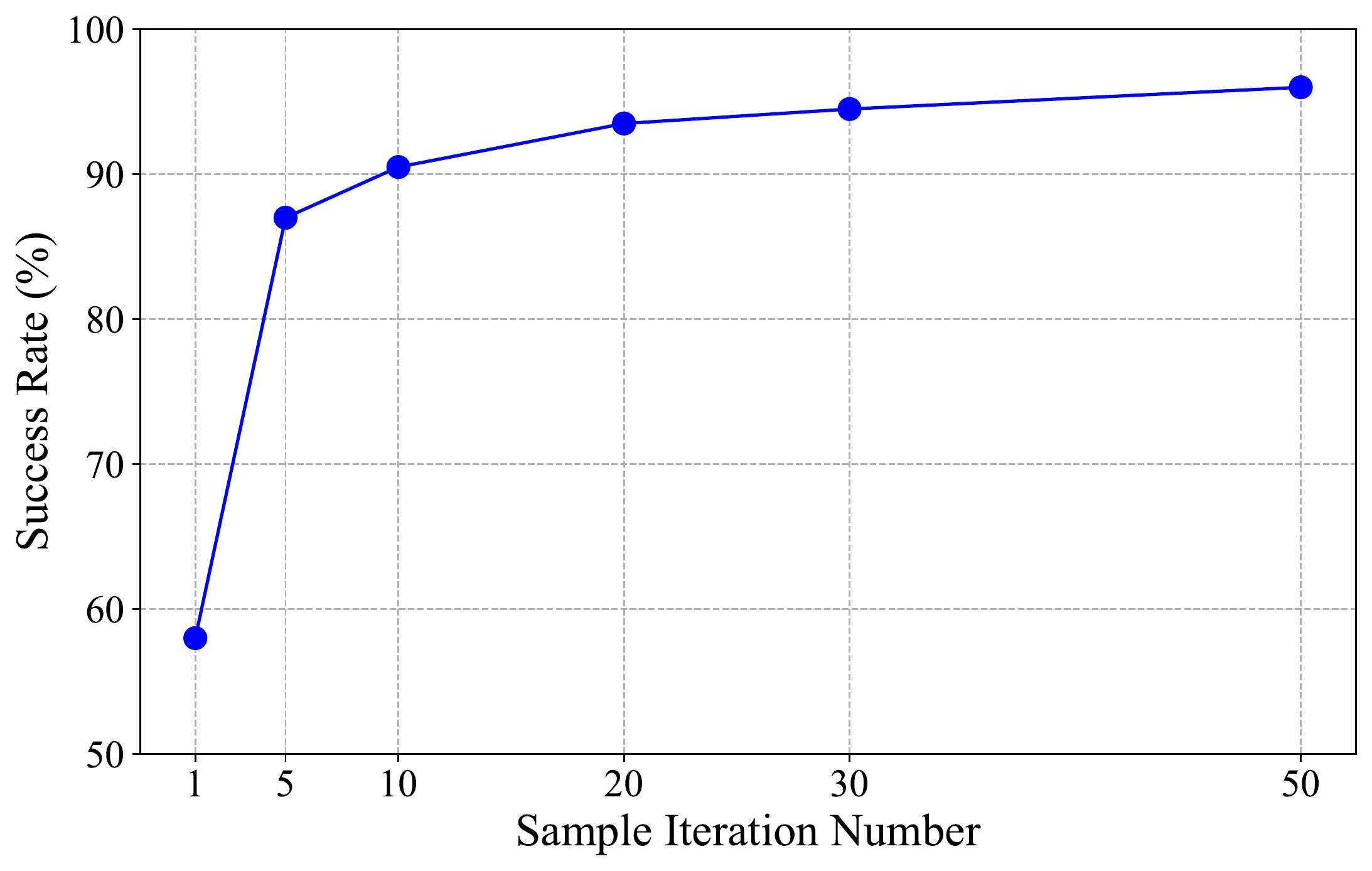}
        \label{fig:anlsBaby_sample}
    \end{minipage}
    \begin{minipage}{0.40\textwidth}
        \centering
        \includegraphics[width=\textwidth]{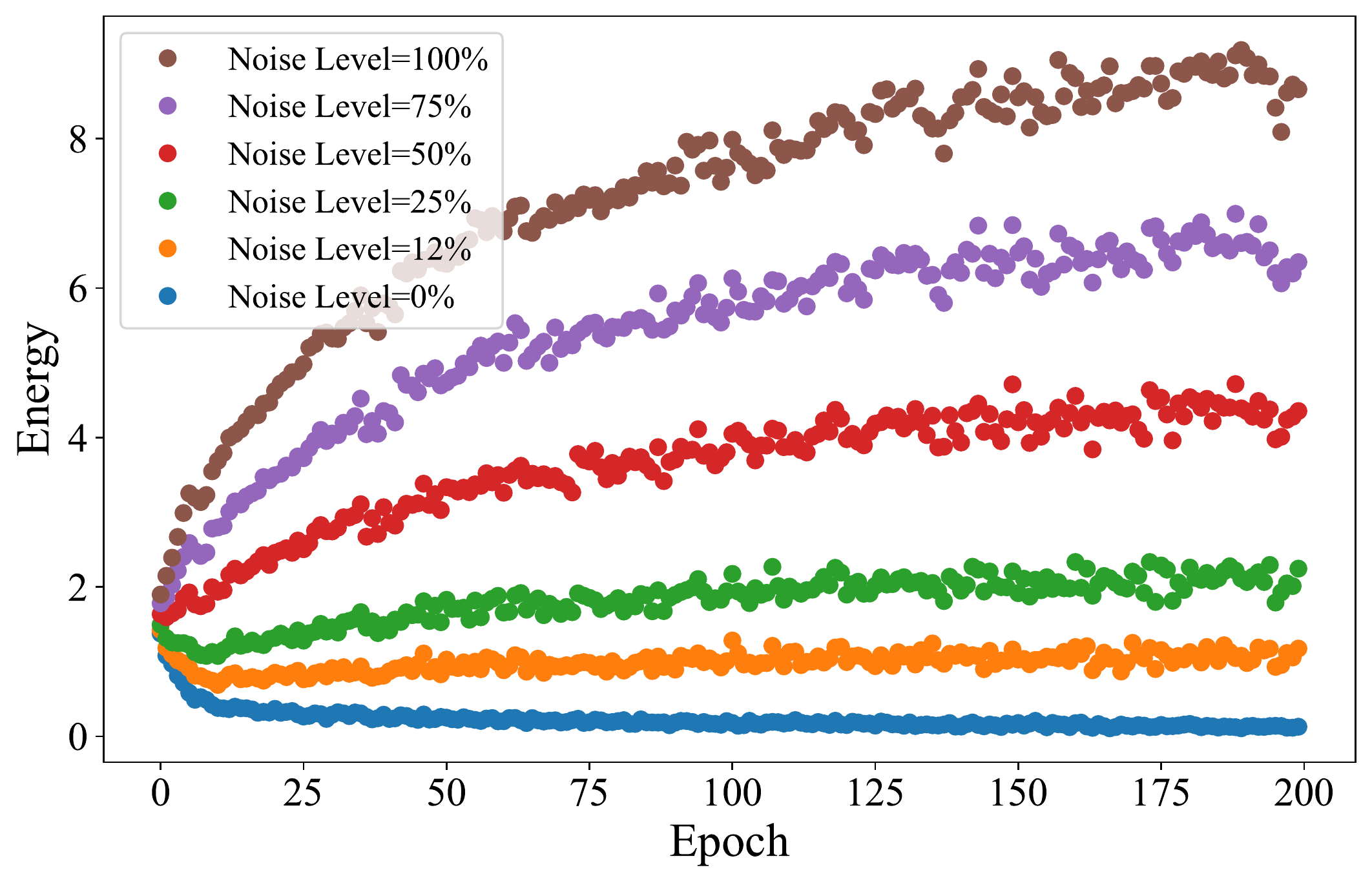}
        \label{fig:anlsBaby_noise}
    \end{minipage}

    \vspace{-20pt}
    \caption{\small
        \textbf{Analysis of \netName in the BabyAI Environments}.
        \textit{Left}: Success rate increases with more sampling steps, suggesting the importance of iterative refinement in \netName.
        \small \textit{Right}: \netName captures the correct energy landscape.
        It assigns low energy to the optimal trajectory (noise level=0\%) and high energy to noisy paths.
    }
    \label{fig:anlsBaby}
    \vspace{-15pt}
\end{figure*}

\myparagraph{Effect of Iterative Refinement.} 
We investigate the effect of iterative refinement by testing the  success  rate of our approach under different sample iteration number in GoToLocalS7N5 environment. 
From the left side of Figure~\ref{fig:anlsBaby}, the task success rate continues to
improve as we increase the number of sample iteration.  

\myparagraph{Energy Landscape Verification.}
We further verify our approach by visualizing the energy   
 assignment on various trajectories in the same environment as above. 
More specifically, we compare the estimated energy of the labeled optimal trajectory with the noisiness of trajectories,
produced by randomizing a percentage of steps in the optimal action sequence.
The right Figure~\ref{fig:anlsBaby} depicts the energy assignment to trajectories  with various corruption levels as a function of training  time.
With the progress of training, \netName learns to
(a) reduce the energy value assigned to the optimal trajectory;
(b) increase the energy value assigned to the corrupted trajectories.
\begin{wraptable}{r}{0.4\textwidth}
    \vspace{-5pt}
    \small
    \centering
  \scalebox{0.68}{
    \begin{tabular}{@{}ccccc@{}}
    \toprule
     & \multicolumn{2}{c}{\bf Optimal} & \multicolumn{2}{c}{\bf Suboptimal(25\%)} \\
    \multicolumn{1}{c}{\bf Env}& {\bf DT} & {\bf \netName} & {\bf DT} & {\bf \netName} \\
     \midrule
    {GoToObjMazeS4}& 46.5\% & \bf{65.0\%} & 45\% & \bf{63.0\%} \\ 
    {GoToObjMazeS7}& 22.0\%&\bf{45.5\%} & 19.5\%&  \bf{44.0\%} \\
    {GoToSeqS5R2} & 26.5\%& \bf{38.0\%} & 29.5\%& \bf{37.5\%} \\
    \bottomrule
    \end{tabular}
  }
  \vspace{-5pt}
 . \caption{\small \textbf{Performance on Suboptimal Data} The task success rate of \netName and DT, using optimal trajectories and suboptimal trajectories containing $25\%$ random actions respectively as training data.}
  \label{tbl:suboptimal_data}
  \vspace{-15pt}
\end{wraptable}
This result justifies the performance of \netName.

\myparagraph{Effect of Training Data.} In BabyAI, we utilize a set of demonstrations generated  using an oracle planner. We further investigate the performance of \netName when the training data is not optimal. To achieve it, we randomly swap the decisions in the optimal demonstration with an arbitrary action with the probability of $25\%$. We compare against DT, the autoregressive sequential planner. Despite a small performance drop in Table~\ref{tbl:suboptimal_data}, \netName still substantially outperforms DT, indicating \netName works well with non-perfect data.

\vspace{-5pt}
\subsection{Atari}

\vspace{3pt}

\myparagraph{Setup} We further evaluate our approach on the dynamically changing Atari environment, with higher-dimensional visual state. 
Due to above features , we train and test our model without the goal state, and update the plan after each step of execution to avoid unexpected changes in the world. We compare our model to BC, DT \citep{chen2021dt}, CQL \citep{kumar2020conservative}, REM \citep{agarwal2020optimistic}, and QR-DQN \citep{dabney2018distributional}.
Following \citet{chen2021dt}, the evaluation is conducted on four Atari games (Breakout, Qbert, Pong, and Seaquest), where $1\%$ of data is used for training. 
Human normalized score is utilized for the performance evaluation.

\myparagraph{Results} 
Table \ref{tbl:atari_main} shows the result comparison.
\netName achieves the best average performance.
Specifically, it achieves better or comparable result in 3 games out of 4,
whereas baselines typically perform poorly in more than one games.

\myparagraph{Energy Landscape Verification}
In Atari environment, the training trajectories are generated by an online DQN agent during training, whose accumulated rewards are varied a lot. \netName is trained to estimate the energy values of trajectories depending on their rewards. In Figure~\ref{fig:anlsAtar}, we visualize the estimated energy of different training trajectories and their corresponding rewards in Breakout and Pong games. We notice that the underlying energy value estimated to a trajectory is well correlated with its reward, with low energy value assigned to high reward trajectory. This justifies the correctness of our trained model and further gives a natural objective to assess the relative of quality planned trajectory. In Qbert and Seaquest games that \netName gets low scores, this negative correlation is not obvious showing that the model is not well-trained.
\begin{table*}[t!]
    \centering
    \small
    \begin{minipage}[t]{0.5\linewidth}
        \centering
        \setlength\tabcolsep{1.5pt} 
        \scalebox{0.76}{
        \begin{tabular}{lrrrrrr}
            \toprule
            \multicolumn{1}{c}{\bf Game} & \multicolumn{1}{c}{\bf \netName} & \multicolumn{1}{c}{\bf DT} & \multicolumn{1}{c}{\bf CQL} & \multicolumn{1}{c}{\bf QR-DQN} & \multicolumn{1}{c}{\bf REM} & \multicolumn{1}{c}{\bf BC} \\ 
            \midrule
            Breakout & $\bf{400.9}\hongyi{\pm 54.0}$& $267.5\hongyi{\pm 56.3}$  & $211.1$ & $17.1$ & $8.9$ & $138.9$ \\ 
            Qbert   &  $19.5\hongyi{\pm 1.6}$ & $15.4\hongyi{\pm 6.6}$ & $\bf{104.2}$ & $0.0$ & $0.0$ & $17.3$ \\
            Pong    & $108.9\hongyi{\pm 1.6}$ & $106.1\hongyi{\pm 4.7}$ & $\bf{111.9}$ & $18.0$ & $0.5$ & $85.2$ \\ 
            Seaquest & $1.3\hongyi{\pm 0.2}$ & $\bf{2.5}\hongyi{\pm 0.2}$ & $1.7$ &  $0.4$ & $0.7$  & $2.1$ \\ 
            \midrule
            Avg & $\bf{132.6}$ & $97.9$ & $107.2$ &  $8.9$ & $2.5$  & $60.9$ \\ 
            \bottomrule
        \end{tabular}
        }
        \caption{\small
            \textbf{Quantitative Comparison on Atari.}
            Gamer-normalized scores for the 1\% DQN-replay Atari dataset \citep{agarwal2020optimistic}. We report the mean \hongyi{and standard error} score across 5 seeds. \netName achieves best averaged scores over 4 games and performs comparably to DT and CQL over all games.}
        \label{tbl:atari_main}
    \end{minipage}
    \hfill
    \begin{minipage}[t]{0.46\linewidth}
        \centering
        \begin{minipage}{0.49\textwidth}
            \centering
            \includegraphics[width=\textwidth]{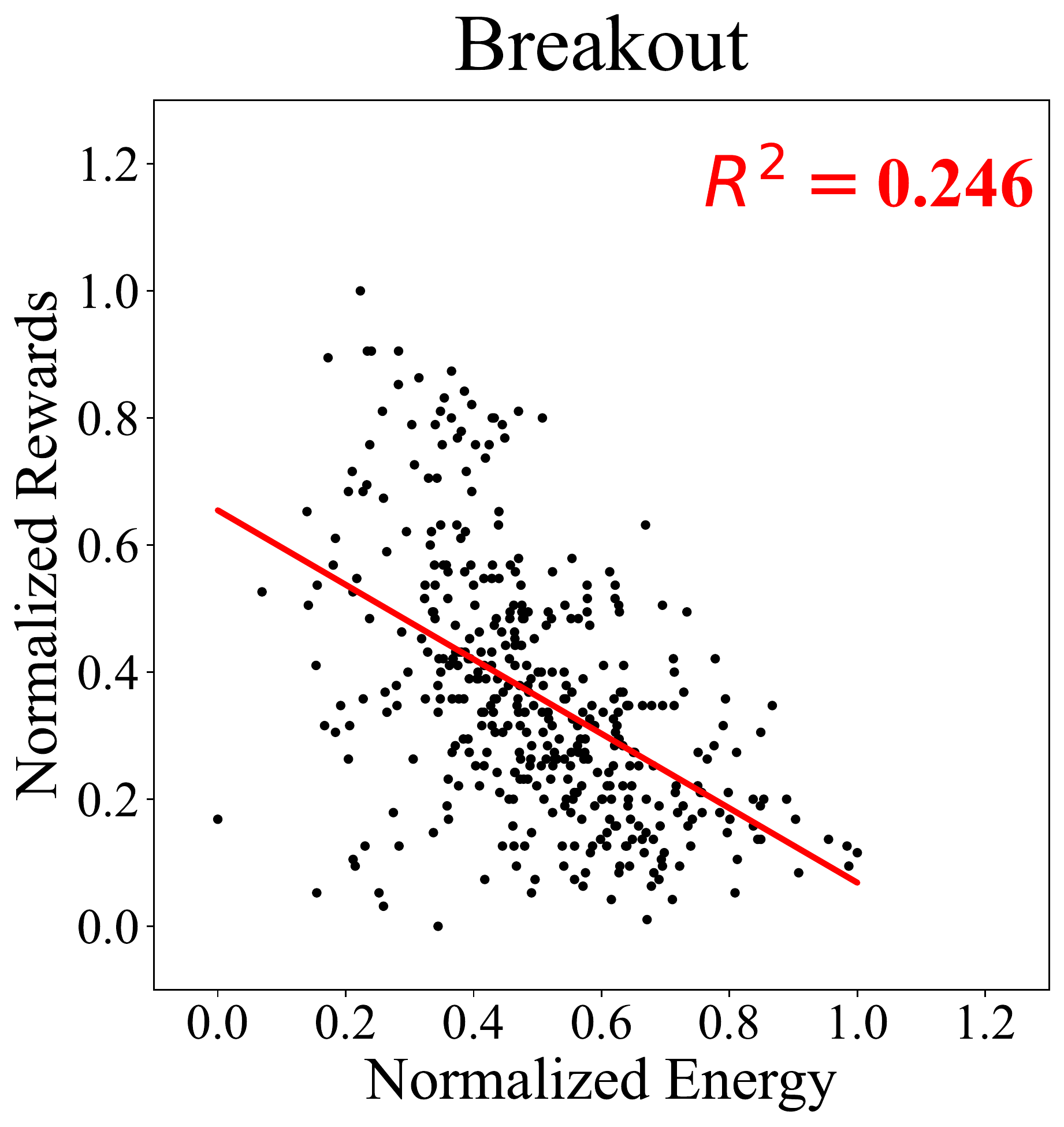}
        \end{minipage}
        \begin{minipage}{0.49\textwidth}
            \centering
            \includegraphics[width=\textwidth]{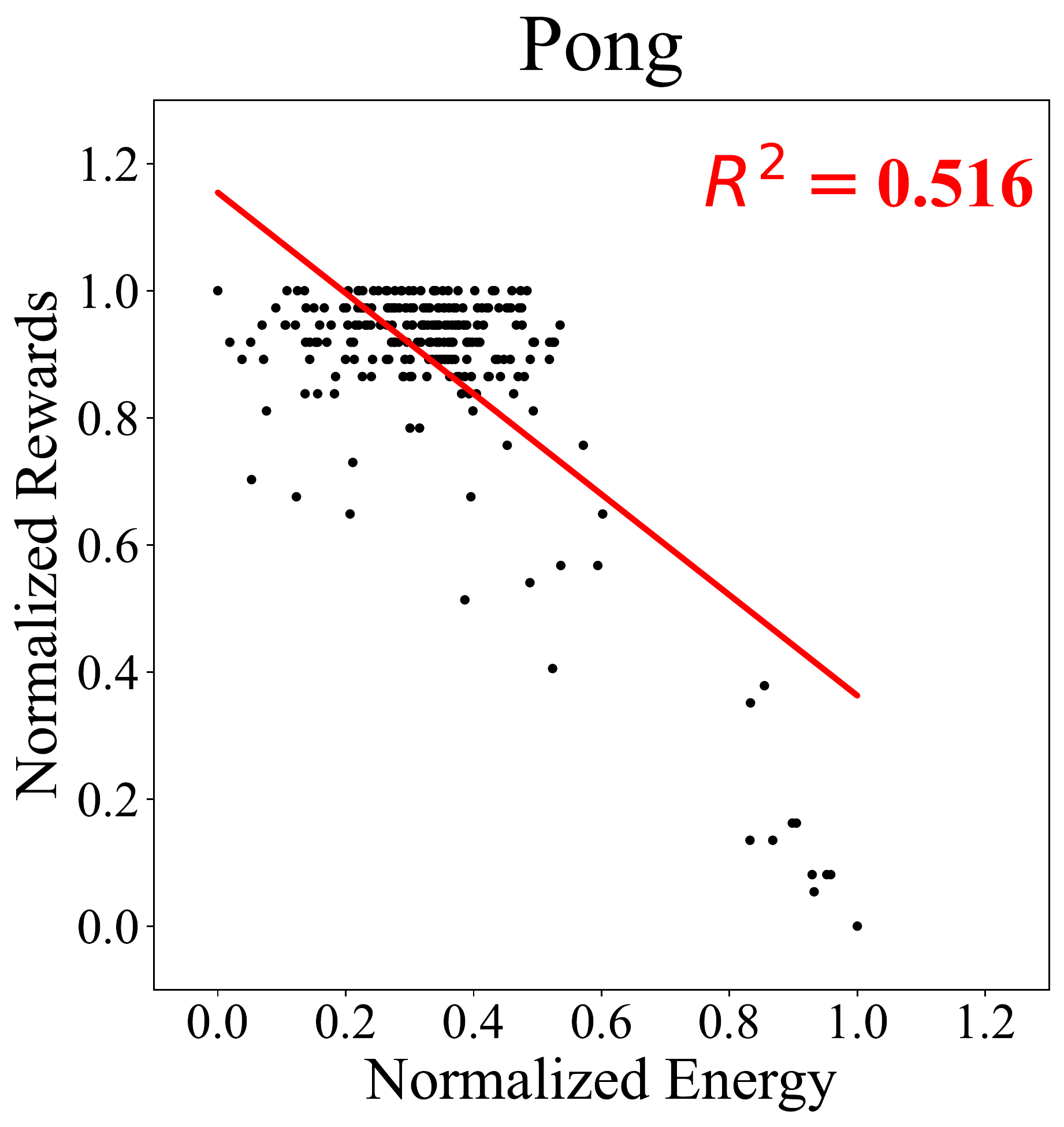}
        \end{minipage}
        \vspace{-10pt}
        \captionof{figure}{ \small 
            \textbf{Energy vs. Reward on Atari}.
            Energies and rewards are normalized to $[0, 1]$.
            We demonstrate negative correspondence between the achieved rewards and estimated energy by \netName,
            which justifies our method.
        }
        \label{fig:anlsAtar}
    \end{minipage}
    \vspace{-18pt}
\end{table*}

\vspace{-7pt}
\section{Properties of Multistep Energy-Minimization Planner}
\vspace{-3pt}

Next, we analyze the unique properties enabled by \netName described in \S \ref{sec:prop} in customized BabyAI environments.
For each environment,
we design at most three settings with increasing difficulty levels to gradually challenge the planning model. 
As before, the target reaching success rate is measured as the evaluation criteria.
The performances are compared against
Implicit Q-Learning (IQL) \citep{kostrikov2021offline} and Decision Transformer (DT) \citep{chen2021dt}.


\vspace{-5pt}
\subsection{Online adaptation} 
\vspace{3pt}

\myparagraph{Setup} To examine \netName's adaptation ability to test-time constraints, 
we construct an BabyAI environment with multiple lethal lava zones at the test time as depicted in Figure~\ref{fig:property} (a).
The planner $\enModel_\theta$ generates the trajectory without the awareness of the lava zones.
Once planned, the energy prediction is corrected by the constraint energy function $\enModel_{\text{constraint}}(\btau{})$, which assigns large energy to the immediate action leading to lava, and zero otherwise.
The agent traverses under the guidance of the updated energy estimation. \hongyi{To make the benchmark comparison fair, we also remove the immediate action that will lead to a lava for all baselines.}
The difficulty levels are characterized by the amount of lava grids added and the way they are added, where \textit{easy}, \textit{medium} correspond to adding at most 2 and 5 lava grids respectively on the way to the goal object in 8$\times$8 grids world. The third case is hard due to the unstructured maze world in which the narrow paths can be easily blocked by lava grids and requires the agents to plan a trajectory to bypass.


\myparagraph{Results}
The quantitative comparison is collected in the Table~\ref{tbl:property_results}, \textit{Left}. 
Although drops with harder challenges, the performance of our model still exceeds both baselines under all settings.
Visual illustration of \textit{medium} example results can be seen in Figure~\ref{fig:property} (a) that the agent goes up first to bypass the lava grids and then drives to the goal object.

\vspace{-3pt}
\subsection{Novel Environment Generalization} 
\vspace{3pt}

\begin{figure*}[t!]
 
  \vspace*{-0.0in}
  \centering
  \scalebox{0.85}{
    \begin{tikzpicture}
     \node[anchor=north west] at (0in,0in)
      {{\includegraphics[width=1.0\textwidth,clip=true,trim=0in
     6.8in 0.3in 0.5in]{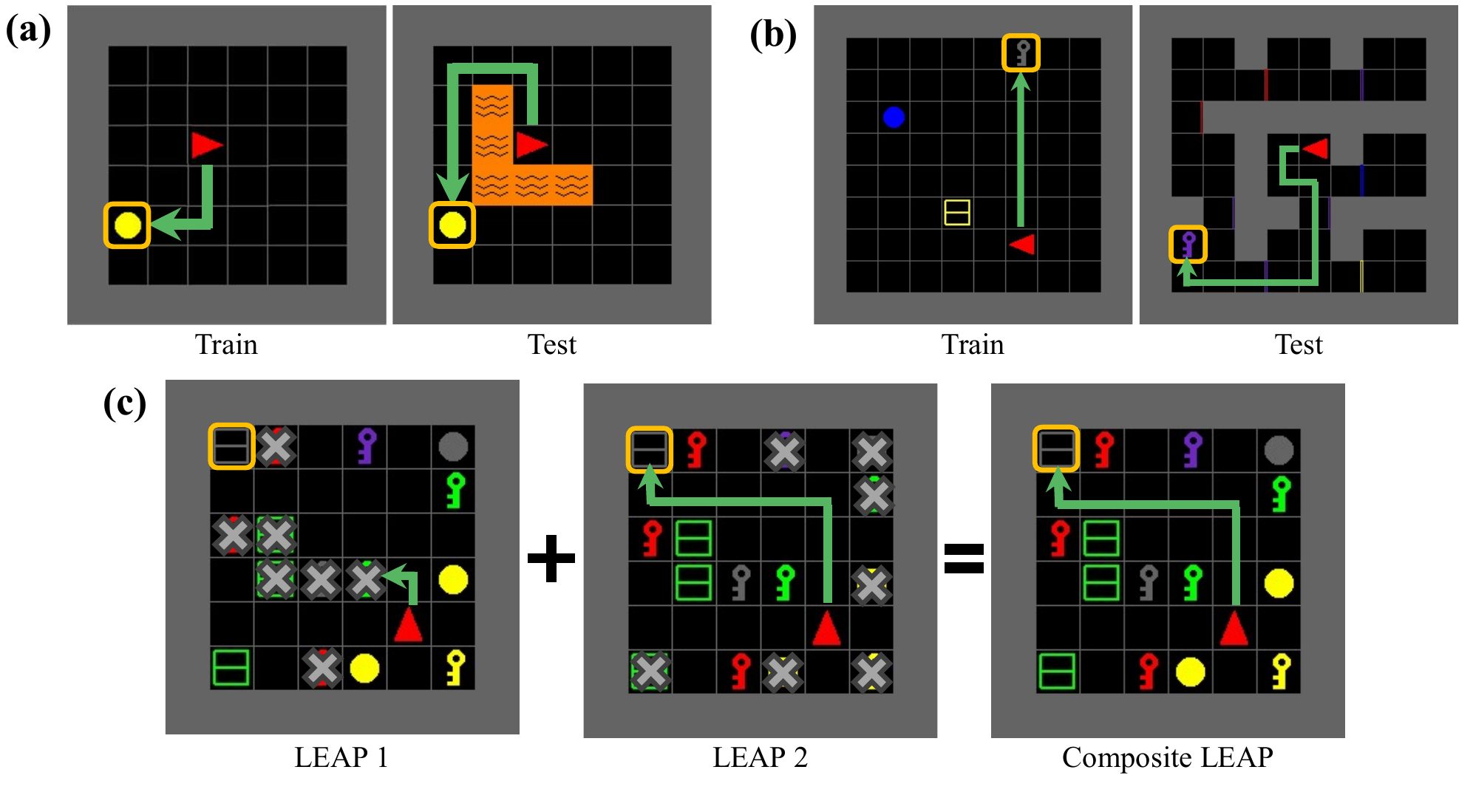}}};
    \end{tikzpicture}
  }
  \vspace{-18pt}
  \caption{\small
    \textbf{Qualitative Visualization of Generalization Tests.} 
    (a): Online adaptation (\textit{medium}), trained in plane world and tested in world with lavas;
    (b): Generalization (\textit{hard}), trained in plane world and tested in maze world;
    (c): Task composition (\textit{easy}), each model only perceive half number of obstacles.
   Target locations and unperceivable obstacles are marked with \protect{\raisebox{-.05cm}{\includegraphics[height=.30cm]{figs/target_highlight.png}}} 
   and
   \protect{\raisebox{-.05cm}{\includegraphics[height=.30cm]{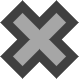}}},
   respectively.
}
  \vspace{-20pt}
 \label{fig:property}
\end{figure*}

\myparagraph{Setup} 
To evaluate \netName's generalization ability in unseen testing environments, we train the model in easier environments but test them in more challenging environments. In \textit{easy} case, the model is trained in 8$\times$8 world without any obstacles but tested in the world with 14 obstacles as distractors. In \textit{medium} and \textit{hard} cases, the model is trained in single-room world but tested in maze world containing multi-rooms connected by narrow paths (Figure~\ref{fig:property} (b)). The maze size and the number of rooms in \textit{hard} case are 10$\times$10 and 9, which are larger than 7$\times$7 and 4 in \textit{medium} case.

\myparagraph{Results}
Our model achieves best averaged performance across three cases, but slightly worse than IQL in \textit{hard} case, see Table~\ref{tbl:property_results}, \textit{Middle}. In contrast, sequential RL model DT has significantly lower performance when moved to in unseen maze environments. \netName trained in plane world could still plan a decent trajectory in unseen maze environment after blocked by walls, see Figure~\ref{fig:property} (b).

\begin{table*}[t!]
\centering
    \small
    \scalebox{0.8}{
    \begin{tabular}{r|rrr|rrr|rrr}
        \toprule
         \multirow{2}{*}{\bf Test} & \multicolumn{3}{c|}{\bf Online Adaptation} & \multicolumn{3}{c|}{\bf Generalization} & \multicolumn{3}{c}{\bf Task Composition} \\
        {} & {\bf \netName} & {\bf DT} & {\bf IQL} & {\bf \netName} & {\bf DT} & {\bf IQL} & {\bf \netName} & {\bf DT} & {\bf IQL}\\
        \midrule
        Easy&$\bf{92.0\%}$& $68.0\%$  & $90.5\%$&$\bf{77.5\%}$ & $36.0\%$ & $60.5\%$&$\bf{83.5\%}$ & $58.0\%$ & $42.5\%$ \\ 
        Medium&$\bf{64.5\%}$& $20.0\%$  & $52.0\%$&$\bf{64.0\%}$ & $37.5\%$ & $57.5\%$&$\bf{43.0\%}$ & $15.5\%$ & $11.5\%$  \\ 
        Hard&$\bf{54.5\%}$&  $48.0\%$ & $44\%$&$61.0\%$ & $24.5\%$ & $\bf{65.5\%}$&N/A&N/A& N/A \\ 
        \bottomrule
    \end{tabular}
    }
    \vspace{-4pt}
    \caption{\small 
    \textbf{Property Test on Modified BabyAI Environments.}
    Three properties performance of \netName and prior algorithms on BabyAI tasks with different difficulty.
    \textit{Left}: Online Adaptation;
    \textit{Middle}: Generalization;
    \textit{Right}: Task Composition.
    }
    \label{tbl:property_results}
    \vspace{-10pt}
\end{table*}

\vspace{-3pt}
\subsection{Task compositionality}
\vspace{3pt}
\myparagraph{Setup}
We design composite trajectory planning and instruction completion tasks for \textit{easy} and \textit{medium} cases respectively. In \textit{easy} case, all obstacles are equally separated into two subsets, each observable by one of the two planners, see Figure~\ref{fig:property} (c). As a result, the planning needs to add up model's predictions using two partial observations to successfully avoid the obstacles. In \textit{medium} case, two separate models trained for different tasks; one for planning trajectory in 10$\times$10 maze world and the other for object pickup in single-room world. The composite task is to complete the object pickup in 10$\times$10 maze world. 

\myparagraph{Results}
Our model significantly outperforms the baselines in both two testing cases, while IQL and DT suffer great success rate drop indicating they can not be applied to composite tasks directly, see Table~\ref{tbl:property_results}, \textit{Right}. This proves that \netName can be easily combined with other models responsible for different tasks, making it more applicable and general for wide-range tasks. In Figure~\ref{fig:property} (c), the composite \netName could reach the goal by avoiding all obstacles even though the first \netName planner is blocked by unperceivable obstacle.

\vspace{-5pt}
\section{Conclusion}
\vspace{-3pt}

This work proposes and evaluates \netName, an \hongyi{sequence} model that plans and refines a trajectory through energy minimization. The energy-minimization is done iteratively - where actions are sequentially along a trajectory. \hongyi{Our current approach is limited to discrete spaces -- relaxing this using approaches such as discrete binning~\citep{janner2021tt} would further be interesting.}

\bibliography{iclr2023_conference}
\bibliographystyle{iclr2023/iclr2023_conference}


\newpage

\appendix
\textbf{\Large{Appendix}}
\vspace{5pt}

\section{Experimental Details}

\subsection{BabyAI Environment Details}
We categorize the environments tested in the trajectory planning and instruction completion into the single-room plane world and multi-room maze world which are connect by doors.
\begin{enumerate}[leftmargin=*]
    \item Trajectory planning: 
    \begin{itemize}
        \item Plane world: GoToLocalS7N5 (7$\times$7), GoToLocalS8N7 (8$\times$8)
        \item Maze world: GoToObjMazeS4 (10$\times$10), GoToObjMazeS7 (19$\times$19)
    \end{itemize}
    \item Instruction completion:
    \begin{itemize}
        \item Plane world: PickUpLoc (8$\times$8)
        \item Maze world: GoToObjMazeS4R2Close (7$\times$7), GoToSeqS5R2 (9$\times$9)
    \end{itemize}
\end{enumerate}

The Table~\ref{tbl:babyai_env_setting} presents the detailed BabyAI environment settings including the environment size, the number of rooms, the number of obstacles, the status of doors and one example instruction in that environment.
\begin{table*}[h]
\centering
\small
\scalebox{0.8}{
\begin{tabular}{lrrrrc}
\toprule
\multicolumn{1}{c}{\bf Env} & \multicolumn{1}{c}{\bf Size} & \multicolumn{1}{c}{\bf \# Room} & \multicolumn{1}{c}{\bf \# Obs} & \multicolumn{1}{c}{\bf Door} & \multicolumn{1}{c}{\bf Inst} \\ 
\midrule
GoToLocalS7N5 & $7\times7$ & 1 & $5$  & $-$ & go to the green key \\ 
GoToLocalS8N7   &  $8\times8$ & 1& $7$ & $-$ & go to the blue box  \\
GoToObjMazeS4    & $10\times10$& 9 & $1$ & $9$ ~$Open$ & go to the blue key \\ 
GoToObjMazeS7 & $19\times19$& 9 & $1$ & $9$ ~$Open$ &  go to the grey ball \\ 
GoToObjMazeS4R2Close & $7\times7$& 4 & $1$ & $3$~ $Closed$ & go to the blue ball \\ 
PickUpLoc   &  $8\times8$& 1 & $8$ & $-$ & pick up the yellow ball \\
GoToSeqS5R2    & $9\times9$& 4 & $4$ & $3$ ~$Closed$ &  \makecell[c]{go to the green door and go to the green door,\\then go to a red door and go to the green door}  \\ 
\bottomrule
\end{tabular}
}
\caption{
BabyAI environment setting details and example instruction.}
\label{tbl:babyai_env_setting}
\end{table*}

\subsection{Network Details} \label{appendix:atari_hyperparameters}

We build our \netName implementation based on Decision Transformer (\url{https://github.com/kzl/decision-transformer}) and exploit the instruction encoder from the BabyAI agent model (\url{https://github.com/mila-iqia/babyai/blob/iclr19/babyai/model.py}). 
In detail, we use the Gated Recurrent Units (GRU) encoder to process the instruction and then apply ExpertControllerFiLM (inspired by FiLMedBlock from \url{https://arxiv.org/abs/1709.07871})to fuse the instruction embedding with state embedding. \hongyi{For all our experiments we use bidirectional mask in transformer attention layer, except for Atari where we found casual attention to perform better.} The full list of hyperparameters can be found in Table \ref{tbl:babyai_hyperparameters} and Table \ref{tbl:atari_hyperparameters}, most of the hyperparameters are taken from Decision Transformer and BabyAI agent model.

\begin{table*}[ht]
\vskip 0.15in
\begin{center}
\begin{small}
\begin{tabular}{ll}
\toprule
\textbf{Hyperparameter} & \textbf{Value}  \\
\midrule
Number of layers & $3$  \\ 
Number of attention heads    & $4$  \\
Embedding dimension    & $128$  \\ 
Batch size   & $64$\\ 
Image Encoder & nn.Conv2d\\
Image Encoder channels & $128, 128$\\
Image Encoder filter sizes & $2 \times 2, 3 \times 3$\\
Image Encoder maxpool strides & $2, 2$ (Image Encoder may vary a little \\ & depending on the environment size)\\
Instruction Encoder & nn.GRU\\
Instruction Encoder channels & $128$\\
State Encoder & nn.Linear\\
State Encoder channels & $256, 256, 128$\\
Max epochs & $200$ \\
Dropout & $0.1$ \\
Learning rate & $6*10^{-4}$ \\
Adam betas & $(0.9, 0.95)$ \\
Grad norm clip & $1.0$ \\
Weight decay & $0.1$ \\
Learning rate decay & Linear warmup and cosine decay (see code for details) \\
\bottomrule
\end{tabular}
\caption{Hyperparameters of \netName for BabyAI experiments.}
\label{tbl:babyai_hyperparameters}
\end{small}
\end{center}
\end{table*}

\begin{table}[ht]
\vskip 0.15in
\begin{center}
\begin{small}
\begin{tabular}{ll}
\toprule
\textbf{Hyperparameter} & \textbf{Value}  \\
\midrule
Number of layers & $6$  \\ 
Number of attention heads    & $8$  \\
Embedding dimension    & $128$  \\ 
Batch size   & $64$ Breakout, Qbert\\ 
            & $128$ Seaquest\\ 
            & $256$ Pong\\ 
Image Encoder & nn.Conv2d\\
Image Encoder channels & $32, 64, 64$\\
Image Encoder filter sizes & $8 \times 8, 4 \times 4, 3 \times 3$\\
Image Encoder strides & $4, 2, 1$ \\
Max epochs & $10$ \\
Dropout & $0.1$ \\
Learning rate & $6*10^{-4}$ \\
Adam betas & $(0.9, 0.95)$ \\
Grad norm clip & $1.0$ \\
Weight decay & $0.1$ \\
Learning rate decay & Linear warmup and cosine decay (see code for details) \\
\bottomrule
\end{tabular}
\end{small}
\caption{Hyperparameters of \netName for Atari experiments.}
\label{tbl:atari_hyperparameters}
\end{center}
\end{table} 

\subsection{Baseline Models}
\paragraph{BabyAI Baseline Models}
We ran BCQ and IQL based on the following implementation
\begin{center}
{
    \small
    \href{https://github.com/sfujim/BCQ}
    {\texttt{https://github.com/sfujim/BCQ}}.
}
\end{center}
\begin{center}
{
    \small
    \href{https://github.com/BY571/Implicit-Q-Learning/tree/main/discrete_iql}
    {\texttt{https://github.com/BY571/Implicit-Q-Learning/tree/main/discrete\_iql}}.
}
\end{center}
For BC and DT, we use the author's original implementation
\begin{center}
{
    \small
    \href{https://github.com/kzl/decision-transformer}
    {\texttt{https://github.com/kzl/decision-transformer}}.
}
\end{center}
\hongyi{For PlaTe, we use the author's original implementation
\begin{center}
{
    \small
    \href{https://github.com/Jiankai-Sun/plate-pytorch}
    {\texttt{https://github.com/Jiankai-Sun/plate-pytorch}}.
}
\end{center}
For MOPO, we use the author's original implementation of dynamic model training and policy learning. For RL policy, we adopt the IQL discussed above.
\begin{center}
{
    \small
    \href{https://github.com/tianheyu927/mopo}
    {\texttt{https://github.com/tianheyu927/mopo}}.
}
\end{center}}
The actor network and policy network of BCQ and IQL use the transformer architecture which is the same as architecture in our model, see details above. The original DT and BC already use the transformer architecture so we didn't change. For all baselines, we add the same instruction encoder and image encoder described above to process instruction and image observations.

\paragraph{Atari Baseline Models} The scores for DT, BC, CQL, QR-DQN, and REM in Table \ref{tbl:atari_main} can be found in \citet{chen2021dt}.

\subsection{Experiment details}\label{app:ExpDetails}
\label{appendix:experiment}

\vspace{3pt}
\myparagraph{BabyAI}

For \netName, the larger size environment requires longer horizon $T$ and correspondingly more sampling iterations $N$. After $N$ iteration, all $T$ planned actions will be executed. For DT model, it's beneficial of using a longer context length in more complex environments as shown in its original paper \citep{chen2021dt}. We list out these parameters for \netName and DT models in Table~\ref{tbl:babyai_experiment_setting}. We didn't use context information in \netName in most BabyAI environments as we expect the iterative planning could generate a correct trajectory based solely on the current state observation. While the GoToSeqS5R2 environment requires go to a sequence of objects in correct order and \netName needs to remember what objects have been visited from the context. During training, we randomly select and mask one action in a trajectory. 

\begin{table*}[h]
\centering
\small
\scalebox{0.8}{
\begin{tabular}{ccccc}
\toprule
& \multicolumn{3}{c}{\bf \netName} & \multicolumn{1}{c}{\bf DT} \\ 
\multicolumn{1}{c}{\bf Env} & \multicolumn{1}{c}{\bf context} & \multicolumn{1}{c}{\bf plan } & \multicolumn{1}{c}{\bf sample iteration} & \multicolumn{1}{c}{\bf context} \\
\midrule
GoToLocalS7N5 & 0 & 5 & 10  & 5 \\ 
GoToLocalS8N7   &  0 & 5 & 10  & 5 \\
GoToObjMazeS4    & 0& 10 & 30 & 10 \\ 
GoToObjMazeS7 & 0& 15 & 50 & 15\\ 
GoToObjMazeS4R2Close & 0& 5 & 10 & 5\\ 
PickUpLoc & 0& 5 & 10 & 5\\ 
GoToSeqS5R2    & 20 & 5 & 10 & 20 \\ 
\bottomrule
\end{tabular}
}
\caption{
BabyAI environment experiment details for \netName and DT.}
\label{tbl:babyai_experiment_setting}
\end{table*}

The input to DT model includes the instruction, state context sequence, action context sequence and return-to-go sequence in which the target reward is set to 1 initially. The input to other baseline models are the same except they use normal reward sequence instead of return-to-go sequence. While \netName only use the instruction, state context sequence and action context sequence.
Inside state sequence, each state $\bs_n$ contains the $[x,y,dir,g_x,g_y]$ meaning the agent's x position, y position, direction and goal object's x position, y position (if the goal location is available).

\myparagraph{Atari}
\vspace{3pt}
In dynamically changing Atari environment, \netName use context information in all four games and only execute the first planned action to avoid the unexpected changes in the world, see details in Table~\ref{tbl:atari_experiment_setting}. During training, we randomly  sample and mask one action in a trajectory. 


\begin{table*}[h!]
    \centering
    \small
    \begin{minipage}[t]{0.45\linewidth}
        \centering\color{black}
        \scalebox{1.0}{
        \begin{tabular}{cccc}
        \toprule
        \multicolumn{1}{c}{\bf Env} & \multicolumn{1}{c}{\bf context} & \multicolumn{1}{c}{\bf plan } & \multicolumn{1}{c}{\bf sample iteration}\\
        \midrule
        Breakout & 25 & 5 & 10  \\ 
        Qbert   &  25 & 5 & 10 \\
        Pong    & 25& 5& 10 \\ 
        Seaquest & 25& 10 & 30\\ 
        \bottomrule
        \end{tabular}
        }
        \caption{
        Atari environment experiment details for \netName.}
        \label{tbl:atari_experiment_setting}
    \end{minipage}
    \hfill
    \begin{minipage}[t]{0.45\linewidth}
        \centering\color{black}
        \scalebox{1.0}{
            \begin{tabular}{ccc}
            \toprule
            \multicolumn{1}{c}{\bf Env} & \multicolumn{1}{c}{\bf w/o return} & \multicolumn{1}{c}{\bf w return}\\
            \midrule
            Breakout & 182.0 & 378.9  \\ 
            Qbert   &  41.0 & 19.6 \\
            Pong    & 100.7& 108.9 \\ 
            Seaquest & 0.5& 1.3\\ 
            \bottomrule
            \end{tabular}
        }
        \captionsetup{labelfont={color=black},font={color=black}}
        \caption{\netName performance in Atari environment with and w/o return input.}
        \label{tbl:LEAP_reward}
    \end{minipage}
    \vspace{-0pt}
\end{table*}

\hongyi{
Note that our approach can easily be conditioned on total reward, by simply concatenating the reward as input in the sequence model. One hypothesis is that when demonstration set contained trajectories of varying quality, taking reward as input following will enable the model to recognize the quality of training trajectories and potentially improve the performance. To further validate the importance of the rewards, we test the \netName with and without return-to-go inputs, which sum of future rewards \cite{chen2021dt}. The results show that the performance degradation without the return-to-go inputs, which is shown in Table \ref{tbl:LEAP_reward}.}

\section{\ycReb{Stochastic Environment Testing}}
\label{appendix:stochastic}
\ycReb{
In this section we demonstrate the possibility of extending our method into stochastic settings.
Although \cite{dtStoch} reveals that planning by sampling from the learnt policy conditioned on desired reward could lead to suboptimal outcome due to the existence of stochastic factors,
our model circumvents the problem by formulating the planning as an optimization problem - we use the Gibbs sampling method to find the trajectory with the lowest energy evaluated by the trained model.
Assuming that the frequency of successful actions dominates in the dataset, our model is trained to assign lower energy to trajectories with higher likelihood of reaching the goal.
Consequently, in the stochastic environments, \netName constructs a sequence of actions that has the best opportunity to accomplish the target.
When executing this plan in a stochastic environment, we may also choose to replan our sequence of actions after each actual action in the environment (to deal with stochasticity of the next state given an action). Note then that this sequence of actions will be optimal in the stochastic environment, as we always choose the action that has the maximum likelihood of reaching the final state. 
Also note that multi-step planning can potentially provide advantage over a simple policy to predict the next action in stochastic environments, as such policy simply assigns probability distribution to the immediate next step without the awareness of future step adjustments facing stochastic factors.
}

\ycReb{
To verify the assumptions, we constructed a stochastic testing in BabyAI environment.
The test is created by adopting a stochastic dynamic model, where the agent fails to execute the turning actions \textit{turn left/right} with $20\%$ chance, and instead performs the remaining actions, including \textit{turn right/left}, \textit{forward}, \textit{pickup}, \textit{drop}, and \textit{open}, with uniform probability.
The remaining settings follow BabyAI experiments detailed in Appendix. \ref{app:ExpDetails}, except that we train models using demonstrations generated with the above dynamic model.
Those training data are noisy in the sense that the actions taken are not optimal, and corrections are required from future actions.
We believe \netName, as a multi-step planner, can learn the above correlations between the consecutive actions. 
We compare \netName with the baseline DT, the results of which is collected in the Table. \ref{tbl:stoch_babyai}.
It can be observed that \netName has a superior performance compared to DT on both tested environments,
which indicates both the possibility of applying our approach in the stochastic settings, and the advantage of multi-step planning when facing stochastic factors. 
}

\begin{table*}[h!]
    \centering\color{black}
    \small
        \setlength\tabcolsep{20pt} 
        \scalebox{1.0}{
            \begin{tabular}{ccc}
            \toprule
            \multicolumn{1}{c}{\bf Env} & \multicolumn{1}{c}{\bf \netName} &  \multicolumn{1}{c}{\bf DT}\\
            \midrule
            GoToObjMazeS4 & \textbf{57.5\%} & 30.8\%   \\ 
            GoToObjMazeS7   &  \textbf{33.3\%} & 28.3\%
            \\
            \bottomrule
            \end{tabular}
        }
        \captionsetup{labelfont={color=black},font={color=black}}
        \caption{Comparison of \netName and DT on stochastic settings}
        \label{tbl:stoch_babyai}
   
    \vspace{-0pt}
\end{table*}

\section{\ycReb{Ablation on Energy model and Optimization method}}

\ycReb{We further ablate on our design choices, including the energy model and sampling methods.
We consider Masked Language Model (MLM) and sequence model classifier as the energy model, and random-shooting, Cross-Entropy Method (CEM), and Gibbs Sampling as the optimization approach. 
All combinations are tested, for which the results are collected in Table~\ref{tbl:samplings}.
We observe that the Gibbs sampling gives the best outcome with MLM model and that defining an energy value using a sequence model classifier doesn't work well in all settings.}

\begin{enumerate}[leftmargin=*]
    \item \hongyi{Sampling methods:} 
    \begin{itemize}
        \item \hongyi{Random-shooting: randomly generated 30 action sequences and pick up the one with lowest estimated energy value.}
        \item \hongyi{Cross-Entropy method: randomly generated 30 action sequences, keep the three best sequences with lowest estimated energy values in each iteration. Then we randomly update one action token in the elite sequences to get 30 new sequences for next iteration. The sequence with lowest energy is selected in the final iteration.}
        \item \hongyi{Gibbs sampling: discussed in the main text.}
    \end{itemize}
    \item \hongyi{Energy models:} 
    \begin{itemize}
        \item \hongyi{Sequence model classifier: LSTM sequence model predicts the scalar energy value given the entire trajectory $\btau{}$, and train the loss between ground truth trajectory energy and estimated energy. The optimal trajectories in Babyai are assigned with lowest energy value 0 and the generates suboptimal trajectories are assigned with higher values depending on the degree of randomness.}
        \item \hongyi{MLM: discussed in the main text.}
    \end{itemize}
\end{enumerate}

\begin{table*}[t]
    \centering
    \small
        \scalebox{1.0}{
            \color{black}\begin{tabular}{cccc}
            \toprule
            \multicolumn{1}{c}{\bf Energy model} & \multicolumn{1}{c}{\bf Random-shooting} &  \multicolumn{1}{c}{\bf CEM} &  \multicolumn{1}{c}{\bf Gibbs sampling}\\
            \midrule
            MLM & 23.3\% & 57.5\%  & \textbf{62.5\%}\\ 
            Classifier & 25.0\% & 12.5\%  & 15.5\% 
            \\
            \bottomrule
            \end{tabular}
        }
        \captionsetup{labelfont={color=black},font={color=black}}
        \caption{Comparison of different combination between energy models and sampling methods}
        \label{tbl:samplings}
    \vspace{-0pt}
\end{table*}

\end{document}